\documentclass[final,1p,times]{elsarticle}

\usepackage{amssymb}
\usepackage{booktabs}
\usepackage{amssymb}
\usepackage{tabularx}
\usepackage[colorlinks]{hyperref}

\biboptions{sort&compress}

\newcolumntype{Y}{>{\raggedright\arraybackslash}X}
\journal{Journal of Biomedical Informatics}

\begin{document}

\begin{frontmatter}



\title{A scoping review on multimodal deep learning in biomedical images and texts}


\author[1]{Zhaoyi Sun}
\ead{zhs4003@med.cornell.edu}

\author[1]{Mingquan Lin}
\ead{mil4012@med.cornell.edu}

\author[2]{Qingqing Zhu}
\ead{qingqing.zhu@nih.gov}

\author[1]{Qianqian Xie}
\ead{qix4002@med.cornell.edu}

\author[1]{Fei Wang}
\ead{few2001@med.cornell.edu}

\author[2]{Zhiyong Lu}
\ead{luzh@ncbi.nlm.nih.gov}

\author[1]{Yifan Peng\corref{cor1}}
\ead{yip4002@med.cornell.edu}

\cortext[cor1]{Corresponding author}

\affiliation[1]{organization={Population Health Sciences},
            addressline={Weill Cornell Medicine}, 
            city={New York},
            postcode={10016}, 
            state={NY},
            country={USA}}

\affiliation[2]{organization={National Center for Biotechnology Information (NCBI), National Library of Medicine (NLM)},
            addressline={National Institutes of Health (NIH)}, 
            city={Bethesda},
            postcode={20894}, 
            state={MD},
            country={USA}}

\begin{abstract}
\textbf{Objective:} Computer-assisted diagnostic and prognostic systems of the future should be capable of simultaneously processing multimodal data. Multimodal deep learning (MDL), which involves the integration of multiple sources of data, such as images and text, has the potential to revolutionize the analysis and interpretation of biomedical data. However, it only caught researchers' attention recently. To this end, there is a critical need to conduct a systematic review on this topic, identify the limitations of current work, and explore future directions.

\noindent
\textbf{Methods:} In this scoping review, we aim to provide a comprehensive overview of the current state of the field and identify key concepts, types of studies, and research gaps with a focus on biomedical images and texts joint learning, mainly because these two were the most commonly available data types in MDL research.

\noindent
\textbf{Result:} This study reviewed the current uses of multimodal deep learning on five tasks: (1) Report generation, (2) Visual question answering, (3) Cross-modal retrieval, (4) Computer-aided diagnosis, and (5) Semantic segmentation.

\noindent
\textbf{Conclusion:} Our results highlight the diverse applications and potential of MDL and suggest directions for future research in the field. We hope our review will facilitate the collaboration of natural language processing (NLP) and medical imaging communities and support the next generation of decision-making and computer-assisted diagnostic system development.
\end{abstract}



\begin{keyword}


Multimodal learning \sep Medical images \sep Clinical notes \sep Scoping review
\end{keyword}

\end{frontmatter}


\section{Introduction}

Multimodal deep learning (MDL), which involves the integration of multiple modalities, such as medical images, unstructured text, and structured Electronic Health Records (EHRs) has gained significant attention in biomedical research~\cite{Huang2020-Fusion}. This approach has been proven to improve the accuracy and efficiency of various tasks in clinical decision-making with imaging and structured EHR (i.e., -omics data, lab test data, demographic data)~\cite{Holste2021-vf, Huang2020-Multimodal, Zhou2021-RadFusion}. The heterogeneous data available to clinicians allows for multiple viewpoints to be considered when making decisions and constructing computer-aided diagnosis and prognosis systems. However, the application of MDL with medical imaging data and unstructured free-text data (i.e., clinical reports) is still in its infancy. The emergence of related research has only recently surfaced. For example, in the field of natural language processing (NLP), pre-trained models, such as Bidirectional Encoder Representations from Transformers (BERT)~\cite{Devlin2018-vb} and Generative Pre-trained Transformer 3 (GPT-3)~\cite{Brown2020-dj}, have garnered world-renowned accomplishments in various downstream tasks. Furthermore, multimodal language models, including Contrastive Language Image Pretraining (CLIP)~\cite{Radford2021-lu} and the more recent KOSMOS-1~\cite{Huang2023-ow}, have demonstrated remarkable performances in addressing general domain tasks. This notable progress has simultaneously facilitated the models' applicability within the medical domain. As a result, we believe it is imperative to comprehensively synthesize the past five years' research on MDL in biomedical images and texts, including an overview of research objectives and methodologies, elucidating development trends, and exploring potential broader clinical applications in the future.

Our review is inspired by several related review articles. \citet{Heiliger2022-lm} provided a comprehensive overview of existing multimodal learning methods and related databases in radiology, proposing a modality-based taxonomy based on the structural and design principles of the model. However, it was method-oriented, which might not facilitate clinicians' comprehension of the development of MDL in the medical field from the standpoint of specific applications. \citet{Cui2022-ro} explored the various fusion strategies employed in disease diagnosis and prognosis. However, the multimodal fusion discussed in these articles primarily included structured data from EHRs, with limited attention to unstructured text. Similarly, numerous systematic reviews have synthesized the employment of multimodal artificial intelligence (AI), machine learning, and the Internet of Medical Things (IoMT) within the realm of biomedicine~\cite{Acosta2022-ov, Kline2022-bu, Muhammad2021-tu}. Nonetheless, these investigations exhibited a notable absence of detailed discussions on implementing multimodal language models in the medical domain.

Additionally, the outstanding achievements of deep learning are accompanied by increasing model complexity and a lack of interpretability of AI models that prevents their applicability to clinical scenarios~\cite{Stiglic2020-gr}. Therefore, it becomes necessary to come up with solutions to address this challenge and move toward more transparent AI. Compared to single-model AI, MDL presents unique challenges as explanations of multimodal data are often separated. For example, there are SHAP values for the EHR and a heatmap for the brain images - a visualization of the brain areas affected. But few visualization/explanation methods integrate the data and results, especially with longitudinal data. While many review studies organize and report challenges and opportunities of explainable AI, however, they do not focus on MDL~\cite{Tjoa2021-dd, Zhang2022-gt, Van_der_Velden2022-wb}.

To our knowledge, our paper represents the first review of multimodal deep learning focusing on medical image and text data, explainability, and human evaluation. Our motivation is to foster the application of multimodal language models in the medical field in a more comprehensible manner. Our target readers include clinicians and computer scientists. Specifically, we aim to provide clinicians with insights into the current performance of various pre-training models on different clinical tasks, as well as opportunities to evaluate model interpretability and contribute to developing new public datasets. Meanwhile, we hope that computer scientists will advance the clinical translation of models by focusing on clinical tasks, recognizing the significance of external validation, and increasing model transparency in the clinical translation process.

The review questions and objectives for this scoping review are as follows: The primary research question is: What is the current state of the literature on MDL in biomedical images and texts? This question will be addressed by exploring the following sub-questions: What databases were utilized in these studies? What were multimodal fusion techniques employed in these studies? Which image and text modalities were incorporated in these studies? What metrics were utilized to evaluate the model's performance in these studies? Did these studies employ external validation? Did these studies explicate the model's interpretability?

The organization of the review is as follows: Section \ref{sec:methods} describes the protocol used in planning and executing this systematic review. Section \ref{sec:results} discusses the research directions of five tasks: report generation, visual question answering, cross-modal retrieval, diagnostic classification, and semantic segmentation. Section \ref{sec:discussion} summarizes the limitations and challenges of the current approaches and highlights future research directions. Lastly, Section \ref{sec:conclusion} concludes the final remarks.

\section{Methods}
\label{sec:methods}

Our scoping review follows the Preferred Reporting Items for Systematic Reviews and Meta-Analyses (PRISMA) guidelines~\cite{Tricco2018-zk}.

\subsection{Eligibility Criteria}

Our scoping review focused on research on multimodal deep learning techniques applied to medical images and unstructured text. The inclusion criteria for our review consisted of English-language articles published between 2018 and 2022, including both conference papers and journal articles. We chose this time frame to capture the most up-to-date research in this rapidly evolving field. Additionally, we refer to relevant preprint articles to ensure we can consider cutting-edge research that has yet to be published in peer-reviewed venues.

\subsection{Information Sources}

A search of multiple databases was carried out, including PubMed\footnote{\url{https://pubmed.ncbi.nlm.nih.gov/}}, the Association for Computing Machinery (ACM) Digital Library\footnote{\url{https://dl.acm.org/}}, the Institute of Electrical and Electronics Engineers (IEEE) Xplore Digital Library\footnote{\url{https://ieeexplore.ieee.org/Xplore/home.jsp}}, Google Scholar\footnote{\url{https://scholar.google.com/}}, and Semantic Scholar\footnote{\url{https://www.semanticscholar.org/}}. The most recent search was executed on January 8, 2023.

\subsection{Search Strategy }

All the studies collected in this research were confined to the medical field. Initially, our search comprised three keyword groups: image modality (e.g., medical images and radiology images), text modality (e.g., text and report), and multimodal fusion learning (e.g., multimodal learning, joint fusion, and contrastive learning). We combined these keywords to carry out the first round of collection across five databases. To ensure the comprehensiveness of the articles collected, we conducted a second round of collection on Google Scholar, by adding a fourth application-oriented keyword group (i.e., report generation, visual question answering, and cross-modal retrieval).

\subsection{Study Selection}
\label{sec:study selection}

Title and abstract screening were conducted independently by two reviewers (ZS and ML). In cases of disagreement, studies were subjected to full-text review, and a consensus was reached through discussions. Subsequently, each article was reviewed and labeled according to the tasks. These tasks encompassed report generation, visual question answering, cross-modal retrieval, diagnostic classification, semantic segmentation, and other related tasks, with the possibility for a single article to correspond to multiple tasks. During the screening and the full-text review stages, we excluded review articles, non-medical articles, poor-quality articles, and unimodal studies (i.e., studies focusing solely on images or text). Articles containing modalities without images or text (e.g., omics data, lab test data, and demographic data) were also excluded.

\subsection{Data Extraction and Synthesis}

In our study, we undertook a systematic analysis of each downstream task. Firstly, we explored commonly used datasets for the task at hand, as well as their primary contents. Secondly, we expounded on the commonly employed multimodal frameworks and development trends of the methodology (e.g., fusion embedding, transformer-based attention models, and contrastive language-image pre-training). Subsequently, we summarized the specific image and text modalities covered in the articles, such as chest X-rays (CXR) and radiology reports. Lastly, we sorted out commonly used evaluation metrics for each downstream task, such as the area under the receiver operating characteristic curve (AUC), F1-score, and bilingual evaluation understudy (BLEU)~\cite{Papineni2002-lr}. Of particular note, we considered whether clinical experts were invited for external validation and explanation of the model's interpretability. We believe this has significant implications for enhancing the accuracy of computer-aided diagnosis and prognosis in the future.

\section{Results}
\label{sec:results}

\subsection{Included Studies and Datasets}

\begin{figure}[hbtp]
    \centering
    \includegraphics[width=.45\textwidth]{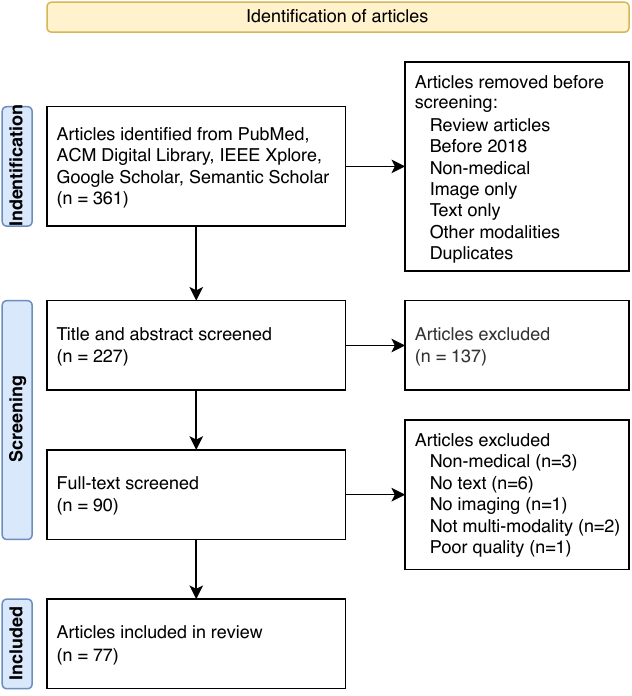}
    \caption{Flowchart of article selection}
    \label{fig:flowchart}
\end{figure}

A total of 361 articles were retrieved from five databases, from which 77 articles were ultimately included in our review. Figure \ref{fig:flowchart} shows the flowchart of our article screening process. During the screening process, we excluded 137 articles based on their titles and abstracts, according to our predetermined exclusion criteria (Section \ref{sec:study selection}). Subsequently, a full-text review was conducted on the remaining articles, which resulted in an additional 13 articles being excluded. Specifically, these articles were discarded based on evaluations of their full texts, including 3 non-medical articles, 6 articles that lacked a text modality, 1 article that lacked an imaging modality, 2 articles on unimodal learning, and 1 poor-quality article.

Table \ref{tab:dataset} encapsulates the medical multimodal datasets employed in the articles collected in this scoping review, encompassing the dataset name, image type, text type, and the corresponding website for each dataset.


\begin{table}[t]
\tiny
\setlength{\tabcolsep}{5pt}
\centering
\caption{Multimodal Medical Image-text Datasets}
\label{tab:dataset}
\begin{tabularx}{\textwidth}{@{}p{10em}p{8em}lY@{}}
\toprule
Dataset & Image type & Text type & URL\\
\midrule
MURA & Bone X-rays & Annotations & \url{https://stanfordmlgroup.github.io/competitions/mura/}\\
DeepLesion & CT & Annotations & \url{https://nihcc.app.box.com/v/DeepLesion}\\
COV-CTR & CT & Radiology reports & \url{https://github.com/mlii0117/COV-CTR}\\
COVID-19 CT & CT & Radiology reports & \url{https://covid19ct.github.io}\\
COVID Rural & CT, CXR & Annotations & \url{https://wiki.cancerimagingarchive.net/pages/viewpage.action?pageId=70226443}\\
COVID-19 Image Data Collection & CT, CXR & Annotations & \url{https://github.com/ieee8023/covid-chestxray-dataset}\\
COVIDx & CXR & Annotations & \url{https://github.com/lindawangg/COVID-Net}\\
MS-CXR & CXR & Annotations & \url{https://aka.ms/ms-cxr}\\
QaTa-COV19 & CXR & Annotations & \url{https://www.kaggle.com/datasets/aysendegerli/qatacov19-dataset}\\
Shenzhen Tuberculosis & CXR & Annotations & \url{https://www.kaggle.com/datasets/raddar/tuberculosis-chest-xrays-shenzhen}\\
SIIM-ACR & CXR & Annotations & \url{https://www.kaggle.com/competitions/siim-acr-pneumothorax-segmentation/data}\\
VinBigData Chest X-ray & CXR & Annotations & \url{https://www.kaggle.com/competitions/vinbigdata-chest-xray-abnormalities-detection/data}\\
RSNA & CXR & Image captions & \url{https://rsna.org/challenge-datasets/2018}\\
CheXpert & CXR & Radiology reports & \url{https://stanfordmlgroup.github.io/competitions/chexpert}\\
IU X-Ray & CXR & Radiology reports & \url{https://openi.nlm.nih.gov}\\
MIMIC-CXR & CXR & Radiology reports & \url{https://physionet.org/content/mimic-cxr/2.0.0}\\
MIMIC-CXR-JPG & CXR & Radiology reports & \url{https://physionet.org/content/mimic-cxr-jpg/2.0.0}\\
NIH-CXR & CXR & Radiology reports & \url{https://nihcc.app.box.com/v/ChestXray-NIHCC}\\
PadChest & CXR & Radiology reports & \url{https://bimcv.cipf.es/bimcv-projects/padchest}\\
RadGraph & CXR & Radiology reports & \url{https://physionet.org/content/radgraph/1.0.0}\\
MoNuSeg & Pathology images & Annotations & \url{https://monuseg.grand-challenge.org/Data}\\
ARCH & Pathology images & Image captions & \url{https://warwick.ac.uk/fac/cross_fac/tia/data/arch}\\
PathVQA & Pathology images & Medical questions & \url{https://github.com/UCSD-AI4H/PathVQA}\\
TCGA & Pathology images & Pathology reports & \url{https://portal.gdc.cancer.gov/repository}\\
PEIR & Pathology images, radiology images & Image captions & \url{https://peir.path.uab.edu/library}\\
MedICaT & Radiology images & Image captions & \url{https://github.com/allenai/medicat}\\
ROCO & Radiology images & Image captions & \url{https://github.com/razorx89/roco-dataset}\\
ImageCLEF VQA-Med & Radiology images & Medical questions & \url{https://www.imageclef.org}\\
SLAKE & Radiology images & Medical questions & \url{https://www.med-vqa.com/slake}\\
VQA-RAD & Radiology images & Medical questions & \url{https://osf.io/89kps}\\
\bottomrule
\end{tabularx}
\end{table}

\subsection{Report Generation}

Report generation aims at generating descriptives from EHR and medical images automatically. It could ease the work burden upon clinicians and improve the quality of the reports themselves. Since the training process of report generation typically requires both medical images and text reports written by clinicians, it can be naturally considered a multimodal learning process.

Table \ref{tab:report generation} provides an overview of the application of multimodal deep learning on report generation.
Common image data used in the medical field include X-rays, computerized tomography (CT), magnetic resonance imaging (MRI), and pathological images. A common dataset for this task is the IU X-Ray~\cite{Demner-Fushman2016-rm} dataset, which comprises 7,470 frontal and lateral chest radiographs and 3,955 corresponding reports. Another widely-used dataset is the MIMIC-CXR~\cite{Johnson2019-et, Johnson2019-gk} dataset, including 377,110 images and 227,827 reports. Furthermore, there exist datasets specifically designed for image classification and assistance in report generation, such as the CheXpert dataset~\cite{Irvin2019-yl}, which comprises 224,316 images and 14 labels marked as present, absent, or uncertain.

Most studies employ convolutional neural networks (CNNs) to process medical images. Regarding text processing, Long Short-Term Memory (LSTM) was previously a popular method. For example, \citet{Yuan2019-ka} developed a CNN encoder and hierarchical LSTM decoder that utilized a visual attention mechanism based on multi-view in radiology. In the recent two years, the Transformer architecture has seen increasing use in report generation. \citet{Chen2022-VMEKNet} proposed the VMEKNet model, which combines the Transformer architecture with visual memory and external knowledge, resulting in improved performance in both qualitative and quantitative experiments and clinical diagnosis. Another notable contribution is the AlignTransformer proposed by \citet{You2021-nb}, which effectively addresses data bias and is particularly well-suited for long-sequence report generation. The use of self-supervised learning techniques, such as CLIP, has also garnered attention for its ability to retrieve reports for report generation purposes. The CXR-RePaiR model proposed by \citet{Endo2021-dm} employed the CLIP approach with retrieval-based mechanisms and achieved outstanding metrics in language generation tasks. Similarly, the RepsNet model proposed by \citet{Tanwani2022-gy} incorporates the principle of self-supervised contrastive alignment. Recent research has focused on improving the factual correctness and completeness of generated reports through reward mechanisms. \citet{Miura2020-ph} developed a model that applies a reward mechanism to reinforcement learning, resulting in significant improvements in clinical performance. This approach was further refined by \citet{Delbrouck2022-Improving} and improved by 14.2\% in factual correctness and 25.3\% in completeness.

%

\begin{table}[t]
\tiny
\centering
\caption{Overview of MDL models for report generation.\label{tab:report generation}}
\begin{tabularx}{\textwidth}{@{}lYYYYYcc@{}}
\toprule
Ref. & Method & Dataset & Image type & Text type & Metrics & External & Explainability \\
 &  &  & & & &  validation &  \\
\midrule
\citet{Yuan2019-ka} & CNN, LSTM & CheXpert, IU X-Ray & CXR & Radiology reports & BLEU, METEOR, ROUGE-L & - & \checkmark\\
\citet{Ni2020-xs} & CNN, LSTM & MIMIC-CXR & CXR & Radiology reports & BLEU, METEOR, ROUGE-L & \checkmark & \checkmark\\
\citet{Nishino2020-lu} & CNN, GRU, BERT & JCT, MIMIC-CXR & CXR & Radiology reports & BLEU, ROUGE, CRS & - & -\\
\citet{Miura2020-ph} & CNN, Transformer & IU X-Ray, MIMIC-CXR & CXR & Radiology reports & BLEU, CIDEr, BERTScore, factENT, factENTNLI & \checkmark & \checkmark\\
\citet{Chen2020-Generating} & CNN, Transformer & IU X-Ray, MIMIC-CXR & CXR & Radiology reports & BLEU, METEOR, ROUGE-L & - & \checkmark\\
\citet{You2021-nb} & CNN, Transformer, multi-head attention & IU X-Ray, MIMIC-CXR & CXR & Radiology reports & BLEU, METEOR, ROUGE-L & \checkmark & -\\
\citet{Alfarghaly2021-wy} & CNN, word2vec, GPT-2 & IU X-Ray & CXR & Radiology reports & BLEU, METEOR, ROUGE-L, CIDEr & \checkmark & \checkmark\\
\citet{Delbrouck2021-zf} & GRU, Fusion & MIMIC-CXR & CXR & Radiology reports & BLEU, METEOR, ROUGE & - & -\\
\citet{Liu2021-Medical} & CNN, BERT, multi-head attention & COVID-19 CT, CX-CHR & CT, CXR & Radiology reports & BLEU, ROUGE-L, CIDEr & \checkmark & \checkmark\\
\citet{Pahwa2021-nx} & CNN, Transformer & IU X-Ray, PEIR Gross & CXR, pathology images & Radiology reports, image captions & BLEU, METEOR, ROUGE-L & - & -\\
\citet{Zhou2021-Visual} & CNN, BioSentVec, LSTM & IU X-Ray, MIMIC-CXR & CXR & Radiology reports & BLEU, METEOR, ROUGE-L, CIDEr, nKTD & - & \checkmark\\
\citet{Endo2021-dm} & CLIP & CheXpert, MIMIC-CXR & CXR & Radiology reports & Semb, BLEU, F1 & - & -\\
\citet{Chen2022-VMEKNet} & CNN, TF-IDF, Transformer & IU X-Ray & CXR & Radiology reports & BLEU, METEOR, ROUGE-L & - & -\\
\citet{Wang2022-ImageSem} & BLIP & ImageCLEF 2020 & Radiological images & Image captions & BLEU, METEOR, ROUGE-L, CIDEr, SPICE, BERTScore & - & -\\
\citet{Yan2022-wl} & CNN, BERT & COV-CTR, IU X-Ray, MIMIC-CXR & CT, CXR & Radiology reports & BLEU, METEOR, ROUGE-L & - & \checkmark\\
\citet{Tanwani2022-gy} & CNN, BERT, BAN & IU X-Ray & CXR & Radiology reports & BLEU & - & \checkmark\\
\citet{Keicher2022-st} & CLIP & MIMIC-CXR & CXR & Radiology reports & AUC & - & -\\
\citet{Chen2022-Cross} & CNN, Transformer, cross-modal memory & IU X-Ray, MIMIC-CXR & CXR & Radiology reports & BLEU, METEOR, ROUGE-L & - & \checkmark\\
\citet{Qin2022-Reinforced} & CNN, Transformer, cross-modal memory & IU X-Ray, MIMIC-CXR & CXR & Radiology reports & BLEU, METEOR, ROUGE-L & \checkmark & \checkmark\\
\citet{Ma2022-ji} & CNN, LSTM, CMCL & IU X-Ray, MIMIC-CXR & CXR & Radiology reports & BLEU, METEOR, ROUGE-L & \checkmark & -\\
\citet{Hassan2022-rb} & CNN, BERT, GRU & IU X-Ray & CXR & Radiology reports & BLEU, ROUGE & - & -\\
\citet{Moon2022-bt} & CNN, BERT, attention masking & IU X-Ray, MIMIC-CXR & CXR & Radiology reports & BLEU, Precision, Recall, F1 & \checkmark & \checkmark\\
\citet{You2022-gl} & CNN, Transformer, GRU & IU X-Ray & CXR & Radiology reports & BLEU, METEOR, ROUGE-L, CIDEr, SPICE, BERTScore & - & \checkmark\\
\citet{Delbrouck2022-Improving} & CNN, BERT, semantic graph-based reward & IU X-Ray, MIMIC-CXR, RadGraph & CXR & Radiology reports & BLEU, ROUGE-L, F1cXb, factENT, factENTNLI, RGE, RGER, RGER & \checkmark & \checkmark\\
\citet{Dalla_Serra2022-tf} & CNN, Transformer & CheXpert, MIMIC-CXR & CXR & Radiology reports & BLEU, METEOR, ROUGE-L & \checkmark & -\\
\bottomrule
\end{tabularx}
\end{table}

Evaluation metrics for report generation can be classified into three categories: text quality, medical correctness, and explainability~\cite{Messina2022-xo}. These metrics are typically intended to be generated automatically, rather than manually, to facilitate automation of the report generation process. The text quality is commonly evaluated using metrics such as BLEU~\cite{Papineni2002-lr}, METEOR~\cite{Banerjee2005-fc}, and ROUGE-L~\cite{Lin2004-vu}. Medical correctness is evaluated using metrics such as AUC, precision, recall, and F1~\cite{Keicher2022-st, Moon2022-bt}. \citet{Yu2022-zd} introduced a composite metric, RadCliQ, aimed at quantifying the similarity between model-generated reports and those produced by radiologists, and the percentage of decreased errors. Additionally, the explainability-related metrics factENT and factENTNLI, proposed by \citet{Miura2020-ph}, have been shown to effectively evaluate the factual correctness and completeness of the model. In the reviewed literature, 10 articles sought external validation through the involvement of radiologists or other clinical experts. Furthermore, 14 articles provided validation of the interpretability of the models through various methods.

\subsection{Visual Question Answering}

In the clinical domain, Visual Question Answering (VQA) represents a computer-assisted diagnostic technique that offers clinical decision-making support for image analysis~\cite{Wu2022-so}.

Table \ref{tab:vqa} is an overview of the application of MDL on VQA. Commencing in 2018, ImageCLEF has been conducting an annual challenge for medical VQA, evaluating and ranking the performance of participating models. The mainstream VQA datasets in the medical domain include VQA-MED-2018~\cite{Hasan2018-lw}, VQA-MED-2019~\cite{Ben_Abacha2019-au}, and VQA-MED-2020~\cite{Ben_Abacha2020-jg}, which were proposed by the challenge tasks. These datasets encompass radiographic images along with corresponding question-answer pairs. For instance, VQA-MED-2020 comprises 4,500 radiographic images and 4,500 question-answer pairs~\cite{Ben_Abacha2020-jg}. Additionally, VQA-RAD consists of 315 radiological images and 3,500 question-answer pairs~\cite{Lau2018-ax}. The PathVQA dataset contains 1,670 pathological images and 32,799 question-answer pairs~\cite{He2020-PathVQA}. \citet{Liu2021-Slake} introduced the SLAKE, a bilingual dataset that encompasses semantic labels and structural medical knowledge, incorporating more modalities and body parts. The SLAKE includes 642 images, 14,028 question-answer pairs, and 5,232 medical knowledge triplets.

A typical VQA model consists of four essential components: an image feature extractor, a question feature extractor, a multimodal fusion component, and a classifier or generator. For the image feature extractor, CNN-based pre-trained models such as ResNet~\cite{He2016-hu} or VGGNet~\cite{Simonyan2014-fa} are often employed to extract high-dimensional features from medical images. \citet{Liu2022-qp} introduced a bi-branch model that leverages both ResNet152 and VGG16 to extract sequence/spatial features and retrieve the similarity of image features, thereby enhancing the semantic understanding of images. For question feature extraction, recurrent neural networks (RNNs) such as Long-Short-Term Memory (LSTM)~\cite{Hochreiter1997-pk} and Gated Recurrent Unit (GRU)~\cite{Chung2014-lq} are commonly utilized. Additionally, BERT-based models~\cite{Devlin2018-vb} have seen increasing use for extracting textual features. With regards to multimodal fusion, models from general domain VQA such as Stacked Attention Networks (SAN)~\cite{Yang2016-cl}, Bilinear Attention Networks (BAN)~\cite{Kim2018-ko}, Multimodal factorized bilinear (MFB)~\cite{Yu2017-pf}, and Multimodal Factorized High-order (MFH)~\cite{Yu2018-ew} are often adopted. \citet{Sharma2021-pd} utilized MFB as a feature fusion technique to design an attention-based model that maximizes learning while minimizing complexity. \citet{Liu2021-Contrastive} proposed a pre-training model called the Contrastive Pre-training and Representation process (CPRD), which effectively resolves the issue of limited MED-VQA data and demonstrates excellent performance.


\begin{table}[t]
\tiny
\centering
\caption{Overview of MDL models for VQA.}
    \label{tab:vqa}
\begin{tabularx}{\textwidth}{@{}lYYYYYcc@{}}
\toprule
Ref. & Method & Dataset & Image type & Text type & Metrics & External & Explainability \\
 &  &  & && &  validation &  \\
\midrule
\citet{Liu2019-fe} & CNN, ETM, MFH & ImageCLEF 2018 & Radiology images & Medical questions & WBSS, BLEU, CBSS & - & -\\
\citet{Ren2020-yp} & CNN, Transformer & ImageCLEF 2019 & Radiology images & Medical questions & Accuracy, BLEU, WBSS & - & -\\
\citet{Zhan2020-bx} & QCR, TCR, MEVF, LSTM, BAN & VQA-RAD & Radiology images & Medical questions & Accuracy & - & -\\
\citet{Liu2021-Contrastive} & CPRD, LSTM, BAN  & SLAKE, VQA-RAD & Radiology images & Medical questions & Accuracy & - & \checkmark\\
\citet{Do2021-pr} & MMQ, LSTM, SAN/BAN  & PathVQA, VQA-RAD & Radiology images, pathology images & Medical questions & Accuracy & - & -\\
\citet{Khare2021-pw} & CNN, BERT, self-attention & ImageCLEF 2019, VQA-RAD & Radiology images & Medical questions & Accuracy & - & \checkmark\\
\citet{Pan2021-rv} & MAML and CDAE, GRU, multi-view attention & VQA-RAD, VQA-RADPh & Radiology images & Medical questions & Accuracy & - & \checkmark\\
\citet{Gong2021-dd} & CNN, LSTM, cross-modal self-attention & VQA-RAD & Radiology images & Medical questions & Accuracy & - & -\\
\citet{Sharma2021-pd} & CNN, BERT, MFB  & ImageCLEF 2019 & Radiology images & Medical questions & Accuracy, AUC-ROC, AUC-PRC & - & \checkmark\\
\citet{Eslami2021-ka} & CLIP, MEVF, QCR & ROCO, SLAKE, VQA-RAD & Radiology images & Medical questions & Accuracy & - & -\\
\citet{Tanwani2022-gy} & CNN, BERT, BAN & VQA-RAD  & Radiology images & Medical questions & Accuracy & - & \checkmark\\
\citet{Chen2022-Multi} & Vision Transformer, BERT, co-attention & ImageCLEF 2019, MedICaT, ROCO, SLAKE, VQA-RAD & Radiology images & Medical questions & Accuracy & - & -\\
\citet{Wang2022-M2FNet} & CDAE, LSTM, attention-based multi-granularity fusion & VQA-RAD & Radiology images & Medical questions & Accuracy & - & \checkmark\\
\citet{Naseem2022-zp} & CNN, LSTM, Transformer & PathVQA  & Radiology images & Medical questions & Accuracy & - & \checkmark\\
\citet{Liu2022-qp} & CNN, Transformer & ImageCLEF 2018, ImageCLEF 2019, VQA-RAD & Radiology images & Medical questions & Accuracy, BLEU & - & -\\
\citet{Haridas2022-ne} & CNN, BERT, ViLBERT & SLAKE & Radiology images & Medical questions & Accuracy & - & -\\
\citet{Moon2022-bt} & CNN, BERT, attention masking & VQA-RAD  & Radiology images & Medical questions & Accuracy & - & \checkmark\\
\citet{Chen2022-Align} & Vision Transformer, BERT, co-attention & ImageCLEF 2019, SLACK, VQA-RAD & Radiology images & Medical questions & Accuracy & - & \checkmark\\
\citet{Pan2022-nd} & MAML, CDAE, GRU, attention-based multimodal alignment & PathVQA, VQA-RAD & Radiology images, pathology images & Medical questions & Accuracy & - & \checkmark\\
\citet{Li2022-Self} & M2I2, Transformer, self-supervised pretraining & ImageCLEF 2022, PathVQA, SLAKE, VQA-RAD & Radiology images, pathology images & Medical questions & Accuracy & - & \checkmark\\
\citet{Zhan2022-gu} & Vision Transformer, BERT, adversarial masking & ROCO, SLAKE, VQA-RAD & Radiology images & Medical questions & Accuracy & - & \checkmark\\
\bottomrule
\end{tabularx}
\end{table}

The issue of data scarcity and lack of multilevel reasoning ability in Med-VQA has prompted the development of the Mixture of Enhanced Visual Features (MEVF)~\cite{Nguyen2019-ap}. MEVF is a meta-learning-based approach that utilizes Model-Agnostic Meta-Learning (MAML)~\cite{Finn2017-gf} and Convolutional Denoising Auto-Encoder (CDAE)~\cite{Masci2011-xz} to effectively address the problem of insufficient data during image feature extraction. The proposed method has gained widespread use in subsequent studies and has been further improved by the introduction of the Question Conditioned Reasoning (QCR) and Type Conditioned Reasoning (TCR) modules by \citet{Zhan2020-bx}, which enhance the model's reasoning ability. \citet{Do2021-pr} have proposed a Multiple Meta-model Quantifying (MMQ) model that achieves remarkable accuracy with the addition of metadata. The latest trends indicate that BERT and attention-based models are currently the most effective and are expected to be the future of VQA models. The RespsNet-10 proposed by \citet{Tanwani2022-gy} achieved an accuracy of 0.804 on the ImageCLEF 2018 and ImageCLEF 2019 datasets. Meanwhile, the study by~\citet{Zhan2022-gu} investigated the contrastive representation learning model UnICLAM with adversarial masking and obtained an accuracy of 0.831 on the SLAKE dataset.

Accuracy is the most widely used evaluation metric for VQA, typically associated with classification models and closed-ended questions. Meanwhile, some generation models designed to tackle open-ended problems may also employ alternative metrics, such as BLEU or WBSS~\cite{Sogancioglu2017-kt}, for evaluation purposes. While 12 articles have demonstrated the interpretability of the models, there has been a lack of studies that have sought to evaluate the results of VQA models from clinicians.

\subsection{Cross-modal Retrieval}

Cross-modal retrieval encompasses two primary types of retrieval: image-to-text retrieval, which involves retrieving associated text for a given image, and text-to-image retrieval, which involves retrieving the associated image for a given text.

Table \ref{tab:retrieval} summarizes an overview of the application of MDL on cross-modal retrieval. In the medical field, cross-modal retrieval tasks frequently involve radiological images and reports, such as those found in MIMIC-CXR~\cite{Johnson2019-gk} and CheXpert~\cite{Irvin2019-yl} datasets. The ROCO dataset, comprising over 81,000 radiology image-text pairs, is also widely employed in cross-modal retrieval tasks~\cite{Pelka2018-ys}. In addition, a small number of pathological captioning datasets exist. One is the ARCH dataset proposed by \citet{Gamper2021-wv}. It comprises 7,579 image and description pairs extracted from medical articles on PubMed and pathology textbooks.


\begin{table}[t]
\tiny
\centering
\caption{Overview of MDL models for cross-modal retrieval.}
    \label{tab:retrieval}
\begin{tabularx}{\textwidth}{@{}lYYYYYcc@{}}
\toprule
Ref. & Method & Dataset & Image type & Text type & Metrics & External & Explainability \\
 &  &  & && &  validation &  \\
\midrule
\citet{Hsu2018-zb} & CNN, TF-IDF, DAN & MIMIC-CXR & CXR & Radiology reports & MRR, nDCG@K & - & -\\
\citet{Lara2020-bt} & CNN, TF-IDF  & TCGA-PRAD & Pathology images & Pathology reports & Precision, MAP, GM-MAP, P@10, P@30 & - & -\\
\citet{Ni2020-xs} & CNN, LSTM & MIMIC-CXR & CXR & Radiology reports & Accuracy, Precision, Recall, BLEU, ROUGE-L, METEOR & \checkmark & \checkmark\\
\citet{Zhang2020-tx} & CNN, CLIP & CheXpert, MIMIC-CXR & CXR & Radiology reports & Precision@K & \checkmark & \checkmark\\
\citet{Wang2021-ee} & Unified transformer & IU X-Ray, MIMIC-CXR, NIH-CXR & CXR & Radiology reports & Precision@K & - & -\\
\citet{Ji2021-vg} & CNN, Transformer & IU X-Ray, MIMIC-CXR & CXR & Radiology reports & Recall@K & - & -\\
\citet{Huang2021-ni} & CNN, BERT, self-attention & CheXpert & CXR & Radiology reports & Precision@K & - & \checkmark\\
\citet{Chen2022-Multi} & Vision Transformer, BERT, co-attention & ROCO & Radiology images & Image captions & Recall@K & - & -\\
\citet{Maleki2022-vh} & Vision Transformer, Text Transformer, self-attention & ARCH & Pathology images & Image captions & Recall@K & - & -\\
\citet{Moon2022-bt} & CNN, BERT, attention masking & IU X-Ray, MIMIC-CXR & CXR & Radiology reports & Hit@K, Recall@K, Precision@K, MRR & - & \checkmark\\
\citet{Chen2022-Align} & Vision Transformer, BERT, co-attention & ROCO & CXR & Radiology reports & Recall@K & - & \checkmark\\
\citet{Wang2022-MedCLIP} & CLIP & CheXpert, COVID, MIMIC-CXR, RSNA & CXR & Radiology reports & Precision@K & - & \checkmark\\
\bottomrule
\end{tabularx}
\end{table}

Most cross-modal retrieval tasks rely on matching image and text features through contrastive learning. This process involves both global and local feature matching, together with attention mechanisms. For example, \citet{Huang2021-ni} introduced GLoRIA which enables cross-modal retrieval through the averaging of global and local similarity metrics. In a separate study, \citet{Chen2022-Multi} developed self-supervised multimodal masked autoencoders, achieving excellent performances for image-to-text retrieval and text-to-image retrieval on the ROCO dataset. \citet{Maleki2022-vh} proposed LILE, a dual attention network that uses Transformers and an additional self-attention loss term to enhance internal features for text retrieval and image retrieval on the ARCH dataset.

Widely used measurements for assessing the performance of cross-modal retrieval are precision@K~\cite{Moon2022-bt, Lara2020-bt, Huang2021-ni} and Recall@K~\cite{Moon2022-bt, Chen2022-Multi, Ji2021-vg, Maleki2022-vh}, which quantify the accuracy of the first K retrieval results. Another commonly used metric is the mean reciprocal rank (MRR)~\cite{Moon2022-bt, Hsu2018-zb}. Out of the 12 studies in our collection, only 2 works incorporated external validation, while 6 studies assessed the interpretability of their model.

\subsection{Computer-aided diagnosis}

MDL-based computer-aided diagnosis (CAD) is the use of generated output from multimodal data as an assisting tool for a clinician to make a diagnosis. Incorporating text modality in this context has been shown to provide supplementary features that can enhance performance in image classification. Currently, research in CAD mainly focuses on utilizing chest X-ray images in conjunction with corresponding radiological reports. It is expected that future pathological datasets will expand this field of research.

Table \ref{tab:cad} summarizes the application of multimodal deep learning on CAD. There exist several commonly employed multimodal fusion strategies, including image-text embedding and contrastive learning. Image-text embedding refers to merging image and text features, which are then trained using supervised learning. For example, \citet{Wang2018-ss} introduced a Text-Image Embedding network (TieNet), which utilized a multi-task CNN-RNN framework and achieved an AUC of over 0.9 in thorax disease classification. In contrast, contrastive learning often involves image-text alignment and self-supervised learning. \citet{Tiu2022-ey} proposed a self-supervised learning framework, CheXzero, which achieved expert-level performance in zero-shot thoracic disease classification without requiring manual labeling. \citet{Monajatipoor2022-td} developed BERTHop, which leverages PixelHop++~\cite{Chen2020-Pixelhop} and VisualBERT~\cite{Li2019-ba} to enable the learning of associations between clinical images and notes. This model achieved an AUC of 0.98 on the IU X-Ray dataset~\cite{Demner-Fushman2016-rm}.


\begin{table}[t]
\tiny
\centering
\caption{Overview of MDL models for computer-aided diagnosis.}
\label{tab:cad}
\begin{tabularx}{\textwidth}{@{}lYYYYYcc@{}}
\toprule
Ref. & Method & Dataset & Image type & Text type & Metrics & External & Explainability \\
 &  &  & && &  validation &  \\
\midrule
\citet{Wang2018-ss} & CNN, LSTM & IU X-Ray, NIH-CXR & CXR & Radiology reports & AUC & - & -\\
\citet{Daniels2019-it} & DNN & IU X-Ray, NIH-CXR & CXR & Radiology reports & AUC, Precision  & - & -\\
\citet{Yan2019-wq} & CNN & DeepLesion & CT & Annotations & AUC, F1 & - & \checkmark\\
\citet{Weng2019-yq} & CNN, BERT, Early fusion & TCGA, TTH & Pathology images & Pathology reports & AUC & - & -\\
\citet{Lara2020-bt} & CNN, TF-IDF  & TCGA-PRAD & Pathology images & Pathology reports & Accuracy & - & -\\
\citet{Chauhan2020-ei} & CNN, BERT & MIMIC-CXR & CXR & Radiology reports & AUC, F1 & \checkmark & \checkmark\\
\citet{Zhang2020-tx} & CNN, CLIP & CheXpert, COVIDx, MURA, RSNA & X-rays & Annotations, radiology report & AUC, Accuracy & \checkmark & \checkmark\\
\citet{Van_Sonsbeek2021-jp} & CNN, BERT & IU X-Ray, MIMIC-CXR & CXR & Radiology reports & AUC & - & \checkmark\\
\citet{Wang2021-ee} & Unified transformer & IU X-Ray, MIMIC-CXR, NIH-CXR & CXR & Radiology reports & AUC & - & -\\
\citet{Ji2021-vg} & CNN, Transformer & IU X-Ray, MIMIC-CXR & CXR & Radiology reports & AUC & - & -\\
\citet{Liao2021-rn} & CNN, BERT & CheXpert, MIMIC-CXR & CXR & Radiology reports & AUC & - & -\\
\citet{Huang2021-ni} & CNN, BERT, self-attention & CheXpert, RSNA & CXR & Radiology reports & AUC, F1 & - & \checkmark\\
\citet{Zheng2021-hr} & CNN, BERT, self-attention & Multimodal COVID-19 Pneumonia Dataset & CT, CXR, ultrasound & Doctor-patient dialogues & AUC, Accuracy, Precision, Sensitivity, Specificity, F1 & - & -\\
\citet{Zhou2021-Generalized} & Vision Transformer, BERT & COVID-19 Image Data Collection, MIMIC-CXR, NIH-CXR, Shenzhen Tuberculosis, VinBigData Chest X-ray & CXR & Radiology reports & AUC & \checkmark & \checkmark\\
\citet{Yan2022-wl} & CNN, BERT & COV-CTR, IU X-Ray, MIMIC-CXR & CT, CXR & Radiology reports & AUC & - & \checkmark\\
\citet{Monajatipoor2022-td} & Vision Transformer, BERT & IU X-Ray & CXR & Radiology reports & AUC & - & -\\
\citet{Jacenkow2022-uu} & CNN, BERT & MIMIC-CXR & CXR & Radiology reports & AUC & \checkmark & -\\
\citet{Hassan2022-rb} & CNN, BERT, GRU & IU X-Ray & CXR & Radiology reports & AUC & - & -\\
\citet{Moon2022-bt} & CNN, BERT, attention masking & IU X-Ray, MIMIC-CXR & CXR & Radiology reports & AUC, F1 & - & \checkmark\\
\citet{You2022-gl} & CNN, Transformer, GRU & IU X-Ray & CXR & Radiology reports & Accuracy & - & \checkmark\\
\citet{Chen2022-Align} & Vision Transformer, BERT, co-attention & MedICaT, MELINDA, MIMIC-CXR, ROCO & CXR & Radiology reports & Accuracy & - & \checkmark\\
\citet{Wang2022-Multi} & Vision Transformer, BERT & CheXpert, COVIDx, MIMIC-CXR, RSNA & CXR & Radiology reports & AUC & - & -\\
\citet{Wang2022-MedCLIP} & CLIP & CheXpert, COVID, MIMIC-CXR, RSNA & CXR & Radiology reports & Accuracy & - & \checkmark\\
\citet{Tiu2022-ey} & Vision Transformer, CLIP & CheXpert, MIMIC-CXR & CXR & Radiology reports & AUC, MCC, F1 & \checkmark & \checkmark\\
\bottomrule
\end{tabularx}
\end{table}

Studies on COVID-19 diagnosis have recently been another popular trend. \citet{Zheng2021-hr} designed a multimodal knowledge graph attention embedding framework for diagnosing COVID-19, based on clinical images and doctor-patient dialogues. The proposed model performed better than single modality approaches, with an AUC of 0.99. In addition, the MedCLIP proposed by \citet{Wang2022-MedCLIP} achieved better performance than supervised models for the zero-shot classification task of COVID-related datasets.

The metrics employed to assess the performance of diagnostic classification primarily comprise the AUC and the F1-score. Additionally, the Matthews correlation coefficient (MCC) is utilized to assess the dissimilarity between model and expert classifications~\cite{Tiu2022-ey}. Out of the 24 studies gathered, 4 incorporated external validation, while 11 studies focused on elucidating the interpretability of the model.

\subsection{Semantic Segmentation}

This group of studies investigates the effectiveness of image-text contrastive learning, which involves utilizing semantic segmentation to extract visual features that can be juxtaposed with textual features to facilitate the comprehension of the relationship between images and their corresponding textual descriptions (Table \ref{tab:seg}). Additionally, local alignment assessment in contrastive learning is evaluated using semantic segmentation techniques.

Typical datasets employed for semantic segmentation include SIIM~\cite{siim-zb} and RNSA~\cite{Shih2019-cp}. The SIIM dataset consists of 12,047 chest radiographs, along with corresponding manual annotations. Similarly, the RNSA dataset includes 29,700 frontal view radiographs for evaluating evidence of pneumonia. \citet{Boecking2022-em} have recently proposed the MS-CXR dataset, which comprises 1153 image-sentence pairs with annotated bounding boxes and corresponding phrases validated by radiologists. This dataset covers eight distinct cardiopulmonary radiology findings.

Image-text alignment and local representation learning are commonly used in MDL for semantic segmentation. These techniques can help improve the model's accuracy by enabling it to better understand the spatial relationships between different regions in the image and the relationship between visual and textual information~\cite{Zhao2022-oo}. \citet{Li2022-LViT} proposed LViT, which used medical text annotations to improve the quality of image data and guide the generation of pseudo labels, leading to better segmentation performance. \citet{Muller2022-rz} devised a novel pre-training approach, LoVT, which aimed to specifically address localized medical imaging tasks. Their method exhibited superior performance on 10 out of 18 localized tasks in comparison to commonly employed pre-training techniques.

In all the research studies that we have gathered, Dice~\cite{Crum2006-os} has been utilized as a metric for measuring the similarity between predicted segmentation and ground truth. Additionally, mean intersection over union (mIoU) and contrast-to-noise ratio (CNR) have also been employed. Out of the 5 studies in our collection, no work incorporated external validation, while 2 studies assessed the interpretability of their model.


\begin{table}[t]
\tiny
\centering
\caption{Overview of MDL models for semantic segmentation.}
\label{tab:seg}
\begin{tabularx}{\textwidth}{@{}lYYYYYcc@{}}
\toprule
Ref. & Method & Dataset & Image type & Text type & Metrics & External & Explainability \\
 &  &  & && &  validation &  \\
\midrule
\citet{Huang2021-ni} &  CNN, BERT, self-attention & CheXpert, SIIM-ACR & CXR & Annotations & Dice & - & \checkmark  \\
\citet{Muller2022-rz} &  CNN, BERT, CLIP & COVID Rural, NIH-CXR, Object CXR, RSNA, SIIM-ACR & CXR & Annotations & Dice & - & -  \\
\citet{Boecking2022-em} & CNN, BERT & MIMIC-CXR, MS-CXR, RSNA & CXR & Annotations & Dice, mIoU, CNR & - & - \\
\citet{Li2022-LViT} & CNN, Vision Transformer, BERT & MoNuSeg, QaTa-COV19 & CXR, Pathology images & Annotations & Dice, mIoU & - & \checkmark\\
\citet{Wang2022-Multi} & Vision Transformer, BERT & RNSA, SIIM-ACR & CXR & Annotations & Dice & - & - \\
\bottomrule
\end{tabularx}
\end{table}

\subsection{Other Related Tasks}

During our article collection, we identified several works that, while not fitting into the aforementioned categories, are of considerable importance. These works include studies centered on medical image generation, object detection, multimodal predictive modeling, MDL-related databases, and libraries of pre-training models. \citet{Chambon2022-Adapting} fine-tuned the Stable Diffusion model to generate CXR images with realistic-looking abnormalities by employing domain-specific text prompts. In a separate publication, they introduced RoentGen, a model adept at synthesizing CXR images predicated upon text prompts present in radiological reports, resulting in a 25\% enhancement in the representation capabilities of pneumothorax~\cite{Chambon2022-RoentGen}. \citet{Qin2022-Medical} scrutinized the implementation of pre-trained vision language models (VLM) for medical object detection and devised an approach to incorporate expert medical knowledge and image-specific information within the prompt, thereby augmenting the performance of zero-shot learning. \citet{Lin2021-ya} developed a survival prediction model using radiation reports and images to forecast ICU mortality. This model outperformed traditional single-modal machine learning methods with a higher C-index. \citet{Bai2021-km} designed an interactive VQA system that empowers patients to upload their own multimodal data, choose the appropriate model in the library, and communicate with an AI robot for model evaluation. \citet{Delbrouck2022-ViLMedic} presented ViLMedic, a Vision-and-Language medical library, consisting of over 20 pre-trained models for various downstream tasks. This resource facilitates the real-world clinical translation of these models. \citet{Kovaleva2020-pt} released the first publicly available visual dialog datasets for radiology, highlighting the belief that integrating patients' medical history information would enhance the performance of traditional VQA models. \citet{Li2020-ul} summarized the performance of four pre-trained models for multimodal vision-and-language feature learning and visualized their attention mechanism. Evidenced by these studies, we believe multimodal vision-and-language learning will continue to expand its range of applications in the future, with more related databases and model libraries being established to promote its clinical use.

\section{Discussion}
\label{sec:discussion}

Our scoping review identifies research related to MDL in biomedical images and texts on different downstream tasks, with specific attention to the datasets employed, model methodology, evaluation metrics, external validation, and interpretability. Overall, the evidence suggests that deep learning models on multimodal medical image and text data can potentially improve diagnostic accuracy and clinical decision-making, showing promising results in several medical fields, including oncology, radiology, and pathology. However, our review also reveals challenges related to data imbalance, clinical knowledge, model fairness, and human evaluation. These findings are highly relevant to clinicians, researchers, and computer scientists interested in leveraging recent advances in artificial intelligence and deep learning to improve patient care and health outcomes.

In the realm of MDL, acquiring high-quality annotated data is crucial for the development and evaluation of models, yet several challenges persist in obtaining such datasets like MIMIC-CXR. First, the annotation of medical data is a laborious and time-consuming task that requires domain expertise and specialized tools to ensure accuracy and consistency, particularly when annotating both image and text data. This can result in insufficient annotated samples for certain modalities, leading to imbalanced datasets that adversely affect model performance. To address data scarcity and reduce the burden of expert annotation, multimodal meta-learning, and few-shot learning are poised to remain popular research topics in the medical field~\cite{Keicher2022-st, Wang2022-M2FNet}. Second, the current trend in medical datasets predominantly features radiology images and their accompanying reports, with a limited representation of other imaging modalities such as pathological images, ultrasound, endoscopy, and text modalities such as clinical notes. This limits the broader clinical application of multimodal models. Future work should construct more multimodal datasets for different medical scenarios, and integrate these heterogeneous data into a system to realize multimodal cross-scenario learning. Thirdly, data privacy concerns are pronounced in the medical domain, necessitating the protection of sensitive patient information. However, this often leads to a lack of publicly available datasets, exacerbating the issue of insufficient and unbalanced data. Advocating for open-access initiatives can help address this challenge by enabling researchers to access larger and more diverse datasets for model training and evaluation. In addition, implementing advanced privacy-preserving techniques, such as differential privacy and federated learning, can further alleviate privacy concerns while allowing researchers to utilize medical data (paper: Federated learning and differential privacy for medical image analysis).

Incorporating clinical knowledge into medical NLP has been identified as a major research direction that can enhance the model's performance and broaden its application in clinical practice~\cite{Callahan2020-pt, Roy2021-dt, Hao2020-sm}. However, the current research is limited in terms of the integration of clinical knowledge into MDL models. Incorporating clinical knowledge into the encoding stage can help learn useful visual features, leading to more accurate predictions. Specifically, clinical knowledge can provide insights into specific image features that are more clinically relevant, such as lesions or abnormalities, and guide the model to focus on these features during the encoding process. \citet{Chen2022-VMEKNet} integrated external knowledge into the features of TF-IDF and achieved improved performance in both report generation and diagnostic tasks. Furthermore, clinical knowledge can be particularly beneficial in scenarios with limited or new data, such as COVID-19-related datasets, where overfitting is more likely to occur. \citet{Liu2021-Medical} incorporated external knowledge into the COVID-19 CT report generation task, generating fewer irrelevant words and higher BLEU scores. In addition, \citet{Chen2022-Align} demonstrated that aligning, reasoning, and learning using clinical knowledge could achieve higher accuracy than each approach individually in VQA. Future research could explore more sophisticated ways to integrate clinical knowledge into models, such as knowledge graphs and ontologies. Moreover, researchers could examine how clinical knowledge from diverse sources, such as electronic health records, medical literature, or expert opinions, can be integrated to enhance the models' performance and adaptability. It is also important to assess the clinical relevance and impact of models in real-world clinical settings by conducting clinical trials and involving clinicians and patients in the development and validation process.

Human evaluation is essential for assessing the practicality of the model in real-world clinical scenarios and providing insights into the model's decision-making process. However, human evaluation was not widely employed in the studies we collected. Out of the five downstream tasks covered in this review, report generation incorporated more external validation, as observed in 10 of 25 articles. Notably, no studies were found to introduce external validation for VQA or semantic segmentation tasks. The observed phenomenon could be attributed to the fact that human evaluation is time-consuming and costly~\cite{Ching2018-oi}. Additionally, the absence of standardized protocols for human evaluation of MDL models in medical settings poses a significant challenge to the comparison and generalization of findings across studies~\cite{Cowley2022-xz}. Furthermore, the interdisciplinary collaboration between clinicians and computer scientists can be a formidable obstacle, owing to differences in their respective backgrounds and training that can hinder effective communication and seamless teamwork. Besides, clinicians often have limited availability to engage in such collaborative efforts, while computer scientists may face stringent deadlines for developing and testing models. In the future, there is a need to develop and adopt standardized protocols for the human evaluation of MDL models in medical applications. Moreover, interdisciplinary workshops can help bridge the gap between clinicians and computer scientists and facilitate effective collaboration. Finally, effective automated metrics could provide a more objective and efficient approach to evaluating MDL models.

The fairness and explainability of MDL models also exhibit deficiencies. The absence of interpretability of the models engenders challenges in fostering trust in their predictions, thereby limiting their adoption in clinical practice. The lack of transparency in these``black boxes'' further compounds the issue as it hinders the detection of errors and biases, thereby resulting in potential harm to patients~\cite{Xie2022-js}. Out of the 77 articles we collected, only 35 provided an exposition of the interpretability of the model, leveraging techniques such as heat maps and factual metrics. Among them, the visual interpretation of CNN models, which are based on attention mechanisms, has gained increasing traction in the medical field~\cite{He2020-MediMLP}. However, it is worth noting that a significant number of articles do not explicitly consider the inclusion of interpretability as an improvement, and only a few employ a formal counterfactual evaluation {[}49{]}. Future MDL research endeavors must prioritize the development of interpretable models. Standardized methods are needed for evaluating and quantifying the interpretability of these models. Additionally, it is essential to engage in a continuous dialogue between clinicians, researchers, and computer scientists to ensure that the development of MDL models aligns with the values and needs of the medical community.

While our scoping review provides a comprehensive overview of the current state of MDL in biomedical images and texts, several limitations must be considered. First, our search strategy may have missed some relevant studies, as we focused on a limited set of databases and search terms. Second, we tried to understand the current state of the literature from their downstream tasks and applications. Still, there was a lack of a systematic summary of the methodology, particularly regarding the multimodal fusion strategy. Third, the heterogeneity of the included studies makes it difficult to compare and synthesize the evidence across different domains and contexts. Finally, our scoping review did not include a formal quality assessment of the studies, which may have affected the reliability and validity of the evidence. However, we believe the breadth and depth of the evidence we gathered will provide a robust foundation for future research and improvement.

\section{Conclusion}
\label{sec:conclusion}

In this scoping review, we systematically examined the current state of research on MDL in biomedical images and texts based on various downstream tasks, including report generation, visual question answering, cross-modal retrieval, computer-aided diagnosis, and semantic segmentation. Our findings suggest that MDL can potentially improve diagnostic accuracy and clinical decision-making, but it also poses challenges related to data imbalance, clinical knowledge, human evaluation, and model fairness. We also discussed several areas for further investigation and improvement, such as developing more robust evaluation standards, collaborating with interdisciplinary institutions or individuals, and exploring new data sources and modalities. Our review has important implications for clinicians, researchers, and computer scientists interested in leveraging the latest advances in MDL to improve patient care and health outcomes.

\section*{Acknowledgments}

This work was supported by the National Library of Medicine under Award No. 4R00LM013001, NSF CAREER Award No. 2145640, and Amazon Research Award. This work is also supported by the Intramural Research Program of the National Institutes of Health, National Library of Medicine.



\bibliographystyle{plainnat}
\bibliography{ref}





\end{document}